\documentclass[lettersize,journal]{IEEEtran}
\usepackage{amsmath,amsfonts}
\usepackage{algorithmicx}
\usepackage{algorithm}
\usepackage{array}
\usepackage[caption=false,font=normalsize,labelfont=sf,textfont=sf]{subfig}
\usepackage{textcomp}
\usepackage{stfloats}
\usepackage{url}
\usepackage{verbatim}
\usepackage{graphicx}
\usepackage{cite}
\usepackage{times}
\usepackage{latexsym}
\usepackage[T1]{fontenc}
\usepackage[utf8]{inputenc}
\usepackage{hyperref}       
\usepackage{url}            
\usepackage{booktabs}       
\usepackage{amsfonts}       
\usepackage{nicefrac}       
\usepackage{microtype}      
\usepackage{xcolor}         
\usepackage{tabularx} 
\usepackage{enumerate}
\usepackage{xspace}
\usepackage{amsthm}
\usepackage{color}
\usepackage{amsmath}
\usepackage{float}
\usepackage{algorithm}
\usepackage{algpseudocode}
\usepackage{bm}
\usepackage{multirow}
\usepackage{array}
\usepackage{makecell}
\usepackage{graphicx}
\usepackage{makecell}
\usepackage{soul}
\usepackage{bbding}
\usepackage[normalem]{ulem}
\usepackage{enumitem}
\usepackage{colortbl}
\usepackage{arydshln}
\usepackage{inconsolata}
\usepackage{CJKutf8}
\usepackage[T1]{fontenc}
\usepackage[utf8]{inputenc}

\begin{document}
\begin{CJK*}{UTF8}{gbsn}

\title{Fine-Grained Chinese Hate Speech Understanding: Span-Level Resources, Coded Term Lexicon, and Enhanced Detection Frameworks}

\author{Zewen Bai, Liang Yang, Shengdi Yin, Yuanyuan Sun, Hongfei Lin
\thanks{This work was partially supported by the National Natural Science Foundation of China, No. xxxx.}
\thanks{Zewen~Bai, Liang~Yang, Shengdi~Yin, Yuanyuan~Sun and Hongfei~Lin are with the School of Computer Science and Technology, Dalian University of Technology, Dalian, 116024, China~(E-mail: dlutbzw@mail.dlut.edu.cn; liang@dlut.edu.cn; 20201071390@mail.dlut.edu.cn; syuan@dlut.edu.cn; hflin@dlut.edu.cn).}

\thanks{Manuscript received April 19, 2021; revised August 16, 2021.}}

\markboth{Journal of \LaTeX\ Class Files,~Vol.~14, No.~8, August~2021}%
{Shell \MakeLowercase{\textit{et al.}}: A Sample Article Using IEEEtran.cls for IEEE Journals}

\IEEEpubid{0000--0000/00\$00.00~\copyright~2021 IEEE}

\maketitle

\begin{abstract}
The proliferation of hate speech has inflicted significant societal harm, with its intensity and directionality closely tied to specific targets and arguments. In recent years, numerous machine learning-based methods have been developed to detect hateful comments on online platforms automatically. However, research on Chinese hate speech detection lags behind, and interpretability studies face two major challenges: first, the scarcity of span-level fine-grained annotated datasets limits models’ deep semantic understanding of hate speech; second, insufficient research on identifying and interpreting coded hate speech restricts model explainability in complex real-world scenarios. To address these, we make the following contributions: (i) We introduce the Span-level Target-Aware Toxicity Extraction dataset (STATE ToxiCN), the first span-level Chinese hate speech dataset, and evaluate the hate semantic understanding of existing models using it. (ii) We conduct the first comprehensive study on Chinese coded hate terms, LLMs’ ability to interpret hate semantics. (iii) We propose a method to integrate an annotated lexicon into models, significantly enhancing hate speech detection performance. Our work provides valuable resources and insights to advance the interpretability of Chinese hate speech detection research.
\end{abstract}


\section{Introduction}
\IEEEPARstart{T}{he} rapid development of social media has accelerated the spread of hate speech in cyberspace, posing a significant threat to social harmony and individual rights. Hate speech refers to language that expresses discrimination or incites harm against specific groups or individuals based on factors such as race, religion, gender, or sexual orientation \cite{bilewicz2020hate}. Its destructive nature makes it a pressing global issue \cite{silva2016analyzing}. Natural language processing (NLP) technology has made significant progress in the automatic detection of hate speech \cite{ahn2024sharedcon, alkhamissi-etal-2022-token, pavlopoulos2021semeval, mathew2021hatexplain, zampieri2023target}. However, the black-box nature of pretrained models, especially closed-source large models, often leads to a lack of transparency in detection results. Interpretability is a key factor in assessing the credibility and robustness of models. Not only reveals the decision-making mechanisms, but it also enhances user trust and promotes fair governance. However, current research on the application of interpretability in hate speech detection is insufficient, and more research is needed to address the complexities of the online environment.

Approximately 941 million people, or 12\% of the global population, speak Mandarin Chinese as their first language \cite{eberhard2024ethnologue}. However, research on Chinese hate speech detection lags significantly behind. The phonetic and semantic characteristics of the Chinese language \cite{mair1991chinese} and the diversity of internet slang \cite{toxicloakcn} make it difficult for traditional classification methods to capture complex patterns of hate speech. Moreover, existing Chinese studies focus mainly on optimizing model performance, with insufficient exploration of interpretability, such as semantic understanding. As a result, models struggle in handling implicit hate speech or multi-target contexts. Therefore, there is an urgent need to develop finer-grained datasets and interpretability methods for Chinese hate speech to address complex linguistic phenomena in real-world scenarios.

This task faces unique challenges within the Chinese linguistic environment. Firstly, traditional studies often employ sentence-level annotations \cite{ahn2024sharedcon, alkhamissi-etal-2022-token, toxicn}, labeling entire sentences as hateful or non-hateful. This coarse-grained approach struggles to capture the structure of hate speech, such as targeted offensive expressions or implicit discriminatory intent. The vague classifications of sentence-level annotations hinder the ability of models to understand hate semantics in multitarget, multiargument, or implicit hate contexts. For example，

\begin{itemize}

    \item “这些女的媚外不找点好的，找没进化完全的黑人。” \\     \textit{Translation: These women obsess over foreigners, not choosing decent ones but those unevolved Black people.}

    This example targets both “这些女的”(\textit{These women}) and “黑人”(\textit{Black people}), with arguments of \textit{“obsessing over foreigners”} (denigrating women’s partner choices) and \textit{“unevolved”} (racial discrimination). Sentence-level annotations fail to disentangle its multi-target semantics.

    \item “河南人最勤奋了，扫大街最适合。” \\     \textit{Translation: Henan people are so hardworking, best suited for street sweeping.}

    This example ostensibly praises “河南人”(\textit{Henan people})’s “勤奋”(\textit{hardworking}) nature but sarcastically implies regional prejudice, suggesting suitability for low-skill labor. Sentence-level annotations’ vague classifications lack explainability in such implicit hate scenarios.

\end{itemize}

To address this, we introduce the first span-level Chinese hate speech detection dataset (\textsc{STATE ToxiCN}), utilizing a Target-Argument-Group-Hateful quadruple prediction framework. This framework annotates each post with its target span, argument span, targeted group, and hateful attributes. The dataset supports fine-grained detection tasks, providing richer hate semantic information to enhance model accuracy and interpretability.

Second, the research on identifying and interpreting coded hate speech remains limited. Such speech employs covert expressions, often using phono-graphemic variation terms or hateful semantic terms to evade automated detection. In Chinese, these expressions are particularly complex due to logographic features and contextual dependencies.

\begin{itemize}

    \item “基佬” (jī lǎo): A phonetic variation, where “基” (jī) sounds like “gay” in Cantonese, used as a derogatory term for gay men.
    
    \item “魄” (pò): An orthographic variation, formed by combining “白” (white) and “鬼” (ghost), used as a slur against White communities.
    
    \item “坦克” (tank): A dog-whistle term, used to insult women with larger body types.
    
\end{itemize}

These coded hate terms are prevalent online, using various linguistic strategies to evade detection. Based on their formation patterns, we classify these coded hate terms into two categories: phono-graphemic variation terms (e.g., “基佬” (jī lǎo)”, “魄” (pò)), which involve changes in sound or character forms; and hateful semantic terms (e.g., “坦克” (tank), “小仙女”(little fairy)),  which carry offensive connotations rooted in cultural context or specific community usage. Such terms significantly challenge models’ ability to identify hateful intent, as they often lack explicit markers and conventional explanatory mechanisms, thereby limiting overall explainability. Thus, developing fine-grained annotation resources for coded hate terms is a key challenge in advancing Chinese hate speech detection.

To address the lack of resources for span-level Chinese hate speech research, we introduce a \textbf{S}pan-level \textbf{T}arget-\textbf{A}ware \textbf{T}oxicity \textbf{E}xtraction dataset (\textsc{STATE ToxiCN}), a novel dataset containing 8,029 posts and 9,533 quadruples addressing sexism, racism, regional bias, and anti-LGBTQ sentiments. Using this dataset, we evaluate the performance of LLMs in span-level Chinese hate speech detection. Specifically, we annotate: 1) extraction of the targets and arguments from the post, 2) determination of whether each Target-Argument pair constitutes hate speech, and 3) classification of the targeted groups for hateful Target-Argument pairs. 

Furthermore, we propose and annotate a Chinese hate lexicon to tackle the challenges of identifying coded hate terms. This lexicon encompasses two categories: phono-graphemic variation terms (e.g., homophonic or character-based variations) and hateful semantic terms (e.g., semantically disguised implicit expressions and offensive slang). Through a systematic analysis of these coded hate term categories, we conduct two experiments: one for hateful term identification to evaluate model performance in detecting these terms, and another for hateful term explanation to assess the explainability of models in uncovering their contextual intent. This approach designs an interpretability evaluation method, providing a valuable resource for enhancing the detection capabilities of LLMs for Chinese coded hate terms.

Finally, we introduce a two-stage training framework that integrates knowledge from the annotated Chinese coded hate term lexicon into the task of Chinese hate speech detection. To mitigate label leakage, we select two additional widely adopted Chinese hate speech detection datasets for evaluation. Through experiments, we demonstrate that the lexicon-based knowledge significantly improves the model’s detection performance. The main contributions of this work are summarized as follows:

\begin{itemize}

\item We introduce a span-level target-aware toxicity extraction dataset (\textsc{STATE ToxiCN}), a novel resource with 8,029 posts and 9,533 quadruples, addressing the gap in span-level Chinese hate speech data.

\item We develop a Chinese hate lexicon, categorizing coded hate terms into phono-graphemic variation terms and hateful semantic terms, with contextual annotations to enhance detection.

\item We propose a two-stage training framework integrating lexicon knowledge, validated on two additional Chinese hate speech datasets, enhancing model detection capability and explainability.

\end{itemize}

\section{Related Work}

\begin{table*}[t]
\caption{Comparison of hate speech datasets based on \textit{Platforms}, \textit{Language}, \textit{\#Posts}, span-level annotations (\textit{Span}), inclusion of Target (\textit{Tar.}), Argument (\textit{Arg.}), \textit{Group}, and Lexicon information (\textit{Lex.}).}
	\centering
	\begin{tabular}{lllrccccc}
		\toprule
		\textbf{Work} & \textbf{Platforms} & \textbf{Language} & \textbf{\#Posts} & \textbf{Span} & \textbf{Tar.} & \textbf{Arg.} & \textbf{Group} & \textbf{Lex.} \\
		\midrule
		Davidson et al. \cite{davidson2017automated} & Twitter & English & 24,802 &  &  &  &  & \checkmark \\
		Founta et al. \cite{founta2018large} & Twitter & English & 80,000 &  &  &  &  &  \\
		Toxic Spans \cite{pavlopoulos2021semeval} & Civil Comments & English & 10,629 & \checkmark &  &  &  &  \\
		HateXplain \cite{mathew2021hatexplain} & Twitter, Gab & English & 20,148 & \checkmark &  &  & \checkmark &  \\
		TBO \cite{zampieri2023target} & Twitter & English & 4,673 & \checkmark & \checkmark & \checkmark &  &  \\
		\hdashline
		COLD \cite{deng2022cold} & Zhihu, Weibo & Chinese & 37,480 &  &  &  & \checkmark &  \\
		SWSR \cite{swsr} & Weibo & Chinese & 8,969 &  &  &  & \checkmark & \checkmark \\
		CDial-Bias-Utt \cite{CDial} & Zhihu & Chinese & 13,394 &  &  &  & \checkmark &  \\
		CDial-Bias-Ctx \cite{CDial} & Zhihu & Chinese & 15,013 &  &  &  & \checkmark &  \\
		\textsc{ToxiCN} \cite{toxicn} & Zhihu, Tieba & Chinese & 12,011 &  &  &  & \checkmark & \checkmark \\
		\textsc{ToxiCloakCN} \cite{toxicloakcn} & Zhihu, Tieba & Chinese & 4,582 &  &  &  & \checkmark &  \\
		\hdashline
		\textsc{STATE ToxiCN} (Ours) & Zhihu, Tieba & Chinese & 8,029 & \checkmark & \checkmark & \checkmark & \checkmark & \checkmark \\
		\bottomrule
	\end{tabular}
	\label{dataset_work}%
\end{table*}

\subsection{Chinese Hate Speech Detection}
Hate speech detection is a critical task in NLP that has attracted considerable attention recently. Researchers have increasingly turned to pre-trained models to address this issue \cite{caselli2020hatebert, Detoxify, zhou2021hate, alkhamissi-etal-2022-token, ali2022hate}. To facilitate progress in this field, several datasets tailored to hate speech detection have been developed \cite{waseem2016hateful, davidson2017automated, founta2018large, hartvigsen2022toxigen, ataei2022pars}. Pavlopoulos et al. introduce span-level hate speech detection  \cite{pavlopoulos2021semeval}, while the TBO dataset advances the field by pioneering the extraction of Target-Argument-Harmful triples \cite{zampieri2023target}. However, Chinese hate speech detection remains significantly underdeveloped.

While some Chinese hate speech datasets exist, these efforts remain limited to the post-level. TOCP and TOCAB, derived from Taiwan's PTT platform, focus on detecting profanity and abusive language \cite{tocab}. The Sina Weibo Sexism Review (SWSR) centers on identifying sexism \cite{swsr}. The Chinese Offensive Language Dataset (COLD) categorizes sentences into types such as individual attacks and anti-bias \cite{deng2022cold}. \cite{CDial}Zhou et al. introduces CDial-Bias, the first annotated dataset specifically addressing social bias in Chinese dialogues. Lu et al. present \textsc{ToxiCN}, a dataset encompassing both explicit and implicit toxic language samples \cite{toxicn}. Xiao et al. introduce a dataset to evaluate LLMs' robustness against cloaking perturbations \cite{toxicloakcn}.

Although previous studies have provided valuable corpora, span-level research remains unexplored. Existing sentence-level annotated datasets suffer from ambiguous classifications, limiting their ability to evaluate a model's hate semantic understanding. Our proposed \textsc{STATE ToxiCN} is the first span-level Chinese hate speech dataset, featuring annotated Target-Argument-Hateful-Group quadruples. The comprehensive hate semantic information provided by \textsc{STATE ToxiCN} not only enables performance assessment of models but also serves as a valuable training corpus to support hate speech detection tasks.

\subsection{Coded Hate Speech}

Since the 1970s, coded communication has been studied as a strategy of racist speech, which uses subtle symbols and behaviors to derogate out-groups \cite{mcconahay1976symbolic}. For instance, terms such as "welfare queens" \cite{macedo1999dancing}, initially used to denigrate Black women, were mainstreamed through dissemination by political and media figures, ultimately normalizing extreme speech \cite{kelly2010regulating}. Humor is another common coding method. Research has pointed out that the Ku Klux Klan, for example, has used jokes to convey racist stereotypes and promote violence, while using the pretext of "it's just a joke" to mask their true intent of spreading hate \cite{billig2001humour}.

The prevalence of such coded language presents significant challenges for NLP models. To investigate these weaknesses, various perturbation techniques have been proposed. These methods include methods like inserting emojis \cite{kirk2021hatemoji}, token replacements and insertions \cite{garg-ramakrishnan-2020-bae}, and probability-based greedy replacements \cite{ren2019generating}. For Chinese, researchers have highlighted adversarial attacks such as word perturbation, the use of synonyms, and typos \cite{su2022rocbert}. Xiao et al. proposed a perturbation strategy that utilizes model-driven automatic disturbances for detecting offensive language in Chinese \cite{toxicloakcn}.

Unlike existing lexicons \cite{davidson2017automated, swsr, toxicn}, we provide interpretative annotations and targeted group labels, establishing a comprehensive Chinese coded hate term lexicon. While some studies assess model comprehension through automated perturbation techniques, these indiscriminate replacements significantly disrupt the original text's contextual integrity. Our proposed lexicon, derived entirely from authentic online environments, addresses the challenges of coded hate speech with fine-grained term-level annotations, including commonly targeted groups and detailed term explanations.

\section{Dataset Construction}

\subsection{Overview}

In this section, we introduce the construction process of the \textsc{STATE ToxiCN} dataset and the annotated lexicon of Chinese coded hate terms. First, we describe the data sources and filtering procedures. Next, we detail the annotation process and the measures implemented to ensure high annotation quality. We then examine the Inter-Annotator Agreement (IAA) at different levels of granularity. Finally, we present relevant statistics for the \textsc{STATE ToxiCN} dataset.

\subsection{Data Source and Filtering}

Our dataset construction is based on the post-level Chinese hate speech dataset \textsc{ToxiCN} \cite{toxicn}. We develop a high-quality span-level Chinese hate speech dataset through a meticulous two-stage data filtering and subsequent annotation process. Since span-level annotation requires identifying precise Target-Argument Pairs within sentences, it is crucial to filter out irrelevant or unannotatable data that could compromise dataset quality and model performance. Therefore, we adopted a robust two-stage filtering process combining automation and manual review to ensure high-quality annotations.

\paragraph{Stage 1: Automated Filtering}
The initial automated filtering stage efficiently removes data obviously unsuitable for detailed annotation. Automated scripts were employed to filter out texts based on length:

\begin{description}[leftmargin=0.5cm]
	\setlength{\itemsep}{0.2em}
	
	\item[\textbf{Short texts (fewer than 5 characters)}] These typically lack sufficient information or context for meaningful span annotation.
	
	\item[\textbf{Long texts (more than 500 characters)}] These often contain excessive, irrelevant content or unstructured language, significantly complicating manual annotation.
This automated process effectively reduced unsuitable samples, enhancing overall annotation efficiency by allowing annotators to focus on analytically valuable data.

\end{description}

\paragraph{Stage 2: Manual Filtering}
In the second stage, manual filtering, trained annotators focused on ensuring data met the specific requirements for constructing a high-quality, span-level dataset, particularly for Target-Argument Pairs. During the annotation process, annotators marked "suggested deletion" sample IDs, which were then rigorously reviewed by an expert team. Specifically, we filtered out two main types of data:

\begin{description}[leftmargin=0.5cm]
	\setlength{\itemsep}{0.2em}
	
	\item[\textbf{Corrupted encoding}] Samples with garbled encoding leading to unclear semantics, e.g., \texttt{\_\_xDE05\_\_xD83D\_\_这不扣押金？\_\_} (Translation: "This doesn't deduct a deposit?" with corrupted encoding).
	
	\item[\textbf{Meaningless texts}] Texts clearly lacking identifiable Target and Argument spans, e.g., “一直想去东北看看雪” (Translation: "Want to see the snow in Northeast China").

\end{description}

The manual filtering process followed a structured flow: Annotator $X$ marked a sample for deletion, submitted these IDs, and an expert team then reviewed them, with disputes resolved by majority vote. Overall, our filtering standards balanced data authenticity with annotation reliability. This ensured that the \textsc{STATE ToxiCN} dataset could serve as an effective and trustworthy benchmark by removing samples incapable of yielding high-quality span-level annotations, while still retaining core challenges of Chinese hate speech like complex language structure, implicit expressions, slang usage, and cultural context understanding.

\subsection{Data Annotation}
\subsubsection{Annotation Guidelines}

During the annotation process, we establish guidelines and implement multi-stage quality control to ensure the consistency of the annotations, aiming to label Target-Argument-Hateful-Group quadruples in the posts. First, we develop detailed annotation guidelines, including standards for extracting targets and arguments (Target-Argument Pair), criteria for determining hatefulness (Hateful), and methods for classifying groups (Group):

\begin{description}[leftmargin=0.5cm]
	\setlength{\itemsep}{0.2em}
	
	\item[\textbf{Target-Argument Pair}] A span consists of both the target and its corresponding argument extracted from the post. A single post may contain more than one Target-Argument Pair.
	
	\item[\textbf{Hateful}]   If the Target-Argument Pair explicitly or implicitly conveys harm towards the target or other groups, it is labeled as "\textbf{hateful}"; otherwise, it is labeled as "\textbf{non-hate}."
	
	\item[\textbf{Group}] Building on the Target-Argument Pair, this category identifies specific groups targeted by hateful expressions, with a single pair probably involving multiple groups.

\end{description}

Regarding the annotation of Chinese coded hate term lexicon, we also establish annotation guidelines for identifying, categorizing, and labeling the terms, with a focus on their frequent groups (Group) and contextual explanations (Explanation):

\begin{description}[leftmargin=0.5cm]
	\setlength{\itemsep}{0.2em}
	
	\item[\textbf{Group}] Each coded hate term is categorized by the group it targets, such as sexism, racism, regional bias, anti-LGBTQ, or others. Some terms may target multiple groups.

	\item[\textbf{Explanation}] An explanation of the coded hate term is provided, including its literal meaning, extended meanings, the reasons for hatred towards targeted groups, and common usage patterns.
\end{description}

In preparation for further experimentation, we meticulously annotated and categorized the coded hate terms present within the posts. Specifically, we classified these terms into two main linguistic categories based on their characteristics:

\begin{description}[leftmargin=0.5cm]
	\setlength{\itemsep}{0.2em}
	
	\item[\textbf{Phono-Graphemic Variation Term}] Coded hate terms derived from changes in pronunciation or character forms, such as homophonic substitutions, acronyms, character splitting/combining, and emoji replacements.

	\item[\textbf{Hateful Semantic Term}] This category includes all other types of coded hate terms not covered by phono-graphemic variation terms, such as slang profanities, cultural allusions, and terms originating from subculture communities.
\end{description}

\subsubsection{Mitigating Bias.} 

To mitigate bias, we assemble annotators with diverse backgrounds, including differences in gender, age, ethnicity, region, and educational level. We believe this diversity helps reduce bias from a single perspective and enhances the reliability of the annotation results. All annotators possess linguistic expertise and have undergone systematic training. During the annotation process, we primarily employ regular cross-validation and expert arbitration to ensure the objectivity and consistency of annotation results. Additionally, we maintain an online document to record and update Chinese coded hate terms identified in the posts. This annotated lexicon serves not only as a research resource but also helps to align annotators' standards.

\subsubsection{Annotation Procedure}

\begin{table}[t]
    \caption{Fleiss’ Kappa for different granularities.}
    \centering
        \begin{tabular}{cccccc} 
            \toprule
            \textbf{Target Span} & \textbf{Argument Span} & \textbf{If Hateful} & \textbf{Targeted Group} & \\
            \midrule
            0.65 & 0.61 & 0.68 & 0.75 \\
            \bottomrule
        \end{tabular}
    \label{kappa}
\end{table}

\paragraph{Cross-validation and Expert Arbitration.} Each text is independently annotated by at least two annotators, who followed a unified annotation guideline to extract Target-Argument pairs, determine the hatefulness, and classify the groups. After the initial annotation phase, 20\% of the samples are regularly selected for cross-validation, during which other annotators re-annotate these samples to ensure a consistent understanding of the rules and standards across annotators. This approach allows us to identify and resolve potential biases or discrepancies in the annotations in a timely manner. 

For disputed samples with significant annotation differences, an arbitration team of domain experts reviews the samples, considering the textual context and annotation guidelines to determine the most appropriate annotation. We explore inter-annotator agreement (IAA) on \textsc{STATE ToxiCN}, with Fleiss' kappa scores \cite{fleiss1971measuring} for each hierarchy detailed in Table \ref{kappa}. To mitigate bias, we assemble a diverse group of annotators, encompassing a variety of genders, ages, ethnicities, regions, and educational backgrounds. The statistical information is presented in Table \ref{ann_demo}.

\begin{table}[t]
	\centering
    \caption{Annotators Demographics.}
	\begin{tabular}{cc}
        \toprule
		\textbf{Characteristic} & \textbf{Demographics} \\ \hline
		Gender & 8 male, 8 female \\
		Age & 7 age < 25, 9 age $\geq$ 25 \\
		Race & 10 Asian, 6 others \\
		Region & From 9 different provinces \\
		Education & 5 BD, 6 MD, 5 Ph.D. \\ 
		\bottomrule
	\end{tabular}
	\label{ann_demo}
\end{table}

\paragraph{Lexicon Annotation.} To ensure annotation quality, we maintain a shared online document for recording and updating detailed information on Chinese coded hate terms. This document includes explanations of each coded hate term and the specific groups they typically reference. Team members could add newly discovered coded hate terms, which are then evaluated and annotated by the expert team. This dynamic maintenance ensures a consistent understanding of emerging language and coded hate terms among all team members.

Additionally, the shared online document serves as a knowledge-sharing platform, providing the annotation team with consistent references and concrete examples for handling complex or ambiguous posts. This mechanism improves annotation efficiency and enhances the consistency of the annotations. This is the first Chinese hate lexicon with interpretable annotations. This lexicon provides valuable research resources for span-level Chinese hate speech semantic understanding.

We employ JSON files to store the lexicon. The data structure is defined as Sample = (Term, [Groups], Definition), where Groups indicates the groups that are commonly targeted by the coded hate terms. If a term does not have a specific targeted group, then Groups is set to "others". We present annotated examples from the lexicon to illustrate its application.

\begin{itemize}

    \setlength\itemsep{0.1em}
    \item \textbf{Term}: 默 (Mò) \\
          \textbf{Meaning}: Silence \\
          \textbf{Groups}: Racism \\
          \textbf{Definition}: “默”是“黑犬”的拼字，指黑色的狗，对黑人群体的侮辱性称呼。 \\
          \textbf{Definition in English}: The term “默” (Mò, meaning silence) is a spelling variant of “黑犬” (black dog), an insulting term for the Black community.
    \item \textbf{Term}: 金针菇 \\
          \textbf{Meaning}: Enoki mushroom (metaphor for male genitalia) \\
          \textbf{Groups}: Sexism \\
          \textbf{Definition}: 作为一种贬义隐喻用于描述男性生殖器官，暗指男性性能力不足。 \\
          \textbf{Definition in English}: As a pejorative metaphor used to describe male genitalia, it subtly implies insufficient male sexual capacity.
          
\end{itemize}

\subsection{Data Description}

\textsc{STATE ToxiCN} dataset contains a total of 8,029 annotated posts, among which 4,942 posts include hateful content, accounting for 61.55\%. A total of 9,533 quadruples are annotated, with 6,034 of them containing hateful information, making up 63.60\%. We present the statistical details of \textsc{STATE ToxiCN} in Table \ref{datasetlabel}. Gender, Region, and Race are the three most common group types in the dataset. Additionally, "multi-group" refers to Target-Argument pairs that involve hatred directed at multiple target groups, with a total of 854 such instances, accounting for 8.96\%. In addition, the annotated lexicon includes 830 Chinese coded hate terms collected from real online forums.

\begin{table}[t]
	\caption{Statistics of annotated posts from the \textsc{STATE ToxiCN} dataset, including Target, Argument, Group, and Hatefulness classifications.}
	\label{datasetlabel}
	\centering
		\begin{tabular}{llcc}
			\toprule
			\textbf{Category} & \textbf{Subcategory} & \textbf{Count} & \textbf{Percentage (\%)} \\
			\midrule
			\multirow{5}{*}{Groups} & Gender       & 1663 & 17.44 \\
			& Race         & 1232 & 12.92 \\
			& Region       & 1323 & 13.88 \\
			& LGBTQ        & 628  & 6.59  \\
			& Others       & 351  & 3.68  \\
			& Multi-group   & 866  & 9.08  \\
			\midrule
			\multirow{2}{*}{Hateful}     & Hate         & 6063 & 63.60 \\
			& Non-Hate     & 3470 & 36.40 \\
			\midrule
			\textbf{Total}                  & -            & 9533 & 100.00 \\
			\bottomrule
		\end{tabular}
\end{table}

\section{Probing Protocol}

In this section, we design an evaluation protocol consisting of three tasks to assess the hate semantic understanding capability of LLMs.

\paragraph{Task 1: Hate Quadruple Identification}
\label{task1}
This task requires the LLM to extract Target-Argument-Hateful-Group quadruples from a given text $T$.
\begin{quote}

\small
\textbf{Input Text}: 这些女的媚外不找点好的，找没进化完全的黑人。 \\
\textbf{Input Text Translation:} These women obsess over foreigners, not choosing decent ones but those unevolved Black people. \\
\textbf{Model Output}: 这些女的 | 媚外 | Sexism | hate [SEP] 黑人 | 没进化完全 | Racism | hate \\
\textbf{Model Output Translation:} These women | obsess over foreigners | Sexism | hate [SEP] Black people | unevolved | Racism | hate
\end{quote}

\paragraph{Task 2: Coded Hate Term Identification}
\label{task2}
This task requires the LLM to extract the coded hate term from a given text $T$.
\begin{quote}

\small
\textbf{Input Text}: 有很多中国女拳友，还是脑子不好使那种。 \\
\textbf{Input Text Translation:} There are many Chinese female fist friends, or the kind with poor brains. \\
\textbf{Model Output}: 女拳友 \\
\textbf{Model Output Translation:} Female fist friends
\end{quote}

\paragraph{Task 3: Coded Hate Term Explanation}
\label{task3}
This task requires the LLM to explain the meaning of the coded hate term identified in a given text $T$.
\begin{quote}

\small
\textbf{Input Text}: 解释“有很多中国女拳友，还是脑子不好使那种”中“女拳友”的含义。 \\
\textbf{Input Text Translation:} Explain the meaning of "female fist friends" in "There are many Chinese female fist friends, or the kind with poor brains." \\
\textbf{Model Output}: 极端女权群体 \\
\textbf{Model Output Translation:} Extreme feminist group
\end{quote}

\section{Experiment}

\subsection{Baselines}
To evaluate the performance of LLMs with varying parameter sizes in detecting span-level Chinese hate speech, we choose twelve well-known models across three categories:

\noindent \textbf{Open-source Models:} mT5-base \cite{mt5}, Mistral-7B \cite{mistral}, LLaMA3-8B \cite{llama3.3modelcard}, Qwen2.5-7B \cite{qwen2.5}; \textbf{Safety-domain Models:} ShieldLM-14B-Qwen \cite{shieldlm}, ShieldGemma-9B \cite{shieldgemmagenerativeaicontent}, and \textbf{Closed-source LLMs:} LLaMA3.3-70B \cite{llama3.3modelcard}, Qwen2.5-72B \cite{qwen2.5}, Gemini-2.5-Flash \cite{gemini2.5}, Claude-3.7-Sonnet \cite{claude3.7}, GPT-4o \cite{gpt4o}, DeepSeek-v3 \cite{liu2024deepseek}. 

\subsection{Evaluation metrics.}

Due to the ambiguity of Chinese span boundaries, a single evaluation metric may not accurately assess the performance of models in span-level Chinese hate speech detection. To obtain more accurate evaluation results, we therefore utilize both hard and soft matching metrics.

\noindent \textbf{Hard-matching:} A predicted quadruplet, particularly its target and argument components, is deemed correct only if it perfectly matches its corresponding gold label.

\noindent \textbf{Soft-matching:} We adopt the algorithm proposed by Han et al. \cite{soft}, where a prediction is considered correct if the Target and Argument scores achieve a threshold of 0.5.

\paragraph{Task 1: Hate Quadruple Identification} 
We require large language models to extract Target-Argument-Hateful-Group quadruples from the text. To comprehensively evaluate the span-level Chinese hate understanding capability of the models, we analyze the predicted results across hierarchical levels from single elements to quadruples. We adopt Macro-F1 scores as the primary evaluation metrics.

\paragraph{Task 2: Coded Hate Term Identification}
During the experiments, we observed that manually incorporating definitions of coded hate terms can influence model comprehension, thereby introducing biases in the identification process.  Consequently, we require the model to extract all hate terms and match them against the annotated coded hate terms.  We adopt recall scores as the primary evaluation metrics.

\paragraph{Task 3: Coded Hate Term Explanation}
Due to the labor-intensive and time-consuming nature of manually evaluating coded hate term explanations, we randomly sample 100 data points for human annotation. We instruct GPT-4o \cite{gpt4o} to determine the winner between human and LLM explanations and compute the \textbf{Win Rate}, \textbf{Tie Rate}, and \textbf{Loss Rate}. This evaluation approach is widely adopted in the field \cite{qin2023toolllm,li2024translate,xu-etal-2024-good}.

\begin{table}[t]
	\caption{Statistics of train and test datasets, including \textit{\#Posts}, total quads (\textit{Quad.}), hateful quads (\textit{Hateful}), and non-hateful quads (\textit{Non-hate}).}
	\centering
		\begin{tabular}{lrrrr}
			\toprule
			\textbf{Category} & \textbf{\#Posts} & \textbf{Quad.} & \textbf{Hateful} & \textbf{Non-hate} \\
			\midrule
			Train & 6,424 & 7,631 & 4,842 & 2,789 \\
			Test & 1,605  & 1,902 & 1,221 & 681 \\
			\midrule
			\textbf{Total}     & 8,029 & 9,533 & 6,063 & 3,470 \\
			\bottomrule
	\end{tabular}
	\label{train}
\end{table}

\subsection{Experiment Settings}

We fine-tune various Open-source and Safety-domain models for \hyperref[task1]{Task 1}, with the training and testing set sizes detailed in Table \ref{train}. Fine-tuning is conducted using LLaMA-Factory\footnote{https://github.com/hiyouga/LLaMA-Factory} \cite{zheng2024llamafactory} with the LoRA method. Models are initialized with a basic prompt. To prevent overfitting, we monitor training loss trends, setting the training duration to 10 epochs where performance typically stabilizes based on preliminary experiments. To minimize hyperparameter sensitivity and ensure robustness, we explore a range of learning rates, ultimately selecting the one that yielded the highest F1 score on the test set. The final performance is then computed via weighted averaging. All specific hyperparameters used are listed in Table \ref{fin_hyp}.

For evaluating closed-source LLMs, we access their capabilities by calling their respective APIs. In addition to the basic task-defining prompt, these API calls are augmented with a few-shot learning approach, providing one example of hate speech and one example of non-hate speech to guide their responses. For \hyperref[task2]{Tasks 2} and \hyperref[task3]{3}, we use Closed-
source LLM APIs to identify and explain coded hate terms. In \hyperref[task3]{Task 3}, the safety features of GPT-4o \cite{gpt4o} and Claude-3.7-Sonnet \cite{claude3.7} prevent them from responding to our requests, so we only have results for four models. To avoid unfairness in the evaluation, we used GPT-4o \cite{gpt4o} as a judge.

\begin{table}[t]
\caption{Hyperparameter Settings for Model Finetuning.}
\centering
\begin{tabular}{lc}
\toprule
\textbf{Hyperparameters} & \textbf{Value} \\ 
\midrule
Epochs & 10 \\
Batch size & 2 \\
Learning rate & $\{1, 2, 3, 4, 5\} \times 10^{-5}$ \\
Cutoff length & 1024 \\
Compute type & fp16 \\ 
Gradient accumulation & 8 \\
Maximum gradient norm & 1.0 \\ 
\bottomrule
\end{tabular}
\label{fin_hyp}
\end{table}

\section{Results on \textsc{STATE ToxiCN} Dateset}

\begin{table*}[t]
\caption{Performance comparison of various models across different levels of annotated tasks, including \textit{Target}, \textit{Argument}, Target-Argument Pair (\textit{T-A Pair}), Target-Argument-Hateful Triple (\textit{T-A-H Tri.}), and Target-Argument-Hateful-Group Quadruple (\textit{Quad.}) under both Hard and Soft evaluation metrics (in \%).}
	\centering
		\begin{tabular}{lcccccccccc}
			\toprule
			\multirow{2}{*}{\textbf{Model}} 
			& \multicolumn{2}{c}{\textbf{Target}} 
			& \multicolumn{2}{c}{\textbf{Argument}} 
			& \multicolumn{2}{c}{\textbf{T-A Pair}} 
			& \multicolumn{2}{c}{\textbf{T-A-H Tri.}} 
			& \multicolumn{2}{c}{\textbf{Quad.}} \\
			\cmidrule(lr){2-3} \cmidrule(lr){4-5} \cmidrule(lr){6-7} \cmidrule(lr){8-9} \cmidrule(lr){10-11}
			& \textbf{Hard} & \textbf{Soft} & \textbf{Hard} & \textbf{Soft} 
			& \textbf{Hard} & \textbf{Soft} & \textbf{Hard} & \textbf{Soft} 
			& \textbf{Hard} & \textbf{Soft} \\
    		\midrule
			\rowcolor{white} \multicolumn{11}{c}{\textit{Finetuned Models (with Basic Prompt)}} \\
			\midrule
			mT5-base   & 59.15 & 70.55 & 28.63 & 67.03 & 23.33 & 55.90 & 17.76 & 43.34 & 16.60 & 38.61 \\
			Mistral-7B    & 62.97 & 73.69 & \underline{35.58} & \underline{70.90} & 30.55 & 60.49 & 26.15 & 51.01 & \underline{23.72} & 45.62 \\
			LLaMA3-8B     & \textbf{64.07} & 73.74 & \textbf{36.72} & 70.82 & \textbf{31.64} & \underline{60.88} & \textbf{27.04} & 51.62 & \textbf{24.27} & 46.08 \\
			Qwen2.5-7B     & \underline{63.96} & \textbf{74.64} & 35.42 & 70.36 & \underline{30.63} & 60.52 & \underline{26.51} & \textbf{52.86} & 23.70 & \underline{47.03} \\
			\hdashline
			ShieldLM-14B-Qwen       & 63.83 & 73.45 & 34.80 & 70.23 & 30.20 & 59.81 & 26.18 & 51.24 & 23.59 & 45.58 \\  	
			ShieldGemma-9B       & 63.40 & \underline{74.31} & 34.40 & \textbf{71.11} & 29.99 & \textbf{61.51} & 25.64 & \underline{52.70} & 23.49 & \textbf{47.14} \\ 	
    		\midrule
			\rowcolor{white} \multicolumn{11}{c}{\textit{LLM APIs (with Basic Prompt and 2 Examples)}} \\
			\midrule
			LLaMA3.3-70B     & 26.32 & 34.93 & 13.49 & 51.64 & 5.96 & 21.92 & 4.92 & 18.32 & 3.22 & 10.93 \\
			Qwen2.5-72B     & 40.94 & 50.44 & 21.10 & 56.36 & 15.66 & 39.49 & 12.48 & 30.92 & 8.74 & 20.29 \\
			Gemini-2.5-Flash      & 42.81 & 52.50 & 20.34 & 56.10 & 14.92 & 37.85 & 12.15 & 30.30 & 9.69 & 23.43 \\
			Claude-3.7-Sonnet   & 42.77 & 53.17 & 20.15 & 58.26 & 14.92 & 40.06 & 12.21 & 33.26 & 9.27 & 24.80 \\
			GPT-4o          & 46.85 & 58.19 & 22.64 & 62.41 & 17.21 & 46.41 & 13.21 & 35.68 & 9.00 & 23.34 \\
			DeepSeek-v3     & 48.16 & 59.25 & 22.79 & 59.38 & 18.68 & 46.40 & 14.95 & 37.19 & 11.48 & 27.38 \\
			\bottomrule
	\end{tabular}
	\label{tab:comparison}
\end{table*}

\subsection{Can LLMs classify Chinese Target-Argument pairs as hateful and identify the groups?}

\paragraph{Finetuned Models}
In both the triplet and quadruplet annotation tasks, fine-tuned models exhibit a distinct performance edge. For the Target-Argument-Hateful (T-A-H Tri.) task, these models consistently achieve hard-match scores around 26\% and soft-match scores hovering near 52\%. Moving to the more intricate quadruplet task, involving the simultaneous identification of hateful content and target groups, hard-match metrics for fine-tuned models generally range from 25\% to 27\%, with soft-match metrics approximately between 45\% and 47\%. These figures suggest that the primary constraint on their performance in these multi-component extraction tasks is often the precision of target-argument pair identification. Despite modest hard-match results, fine-tuned models generally reach around 50\% in soft-match metrics. Within this group, LLaMA-3-8B demonstrates superior hard-match capabilities, while Qwen2.5-7B and ShieldGemma-9B show stronger performance in soft-match evaluations.

\paragraph{LLM APIs}
Conversely, LLM APIs demonstrate consistently weak performance across both the triplet and quadruplet tasks. In the \textit{T-A-H Tri.} task, LLMs only manage hard-match scores between 6\% and 15\%, and soft-match scores from 20\% to 38\%. The more demanding \textit{Quad.} task sees a further decline in LLM hard-match metrics, falling to just 3\% to 12\%, with soft-match metrics, though marginally better, still limited to 11\% to 27\%. We posit that these performance shortcomings are predominantly a consequence of their challenges in accurately identifying precise span boundaries. These results emphatically illustrate that LLMs, without specialized fine-tuning on Chinese hate speech data, are profoundly inadequate for complex Chinese hate speech triplet and quadruplet predictions. Among the various APIs tested, DeepSeek-v3 stands out with notably better performance, particularly for the \textit{Quad.} task.

\paragraph{Comparison and Summary}
The performance disparity between fine-tuned models and LLM APIs in \textit{T-A-H Tri.} and \textit{Quad.} tasks is stark, with fine-tuning demonstrating a decisive advantage. The primary bottleneck for both model types lies in the identification of \textit{T-A pairs}, as evidenced by the significantly lower hard-match scores for \textit{T-A-H Tri.} and \textit{Quad.} tasks compared to individual span identification. While fine-tuned models, despite this limitation, still achieve soft-match F1 scores approaching 50\%, suggesting a reasonable semantic grasp of the hateful content and target groups, LLM APIs consistently fall short, with their performance plummeting dramatically as task complexity increases.  This indicates that their inherent pre-training for language understanding is insufficient for the multi-span relational extraction required in complex hate speech analysis, even with few-shot prompting.

\definecolor{dec}{RGB}{56,87,35}
\definecolor{rise}{RGB}{192,0,0}
\definecolor{human}{RGB}{47,85,151}

\begin{table*}[t]
\caption{Performance comparison of various models on a coded hate term-containing subset of the test set. Green indicates a \textcolor{dec}{decrease} in F1 score on the subset compared to the full dataset, while red indicates an \textcolor{rise}{increase} (in \%).}
	\label{slang}
	\centering
		\begin{tabular}{lcccccccccc}
			\toprule
			\multirow{2}{*}{\textbf{Model}} 
			& \multicolumn{2}{c}{\textbf{Target}} 
			& \multicolumn{2}{c}{\textbf{Argument}} 
			& \multicolumn{2}{c}{\textbf{T-A Pair}} 
			& \multicolumn{2}{c}{\textbf{T-A-H Tri.}} 
			& \multicolumn{2}{c}{\textbf{Quad.}} \\
			\cmidrule(lr){2-3} \cmidrule(lr){4-5} \cmidrule(lr){6-7} \cmidrule(lr){8-9} \cmidrule(lr){10-11}
			& \textbf{Hard} & \textbf{Soft} & \textbf{Hard} & \textbf{Soft} 
			& \textbf{Hard} & \textbf{Soft} & \textbf{Hard} & \textbf{Soft} 
			& \textbf{Hard} & \textbf{Soft} \\
    		\midrule
			\rowcolor{white} \multicolumn{11}{c}{\textit{Finetuned Models (with Basic Prompt)}} \\
			\midrule
			mT5-base   & 56.83$_{\textcolor{dec}{2.32}}$ & 68.33$_{\textcolor{dec}{2.22}}$ & 27.17$_{\textcolor{dec}{1.46}}$ & 64.17$_{\textcolor{dec}{2.86}}$ & 21.33$_{\textcolor{dec}{2.00}}$ & 51.17$_{\textcolor{dec}{4.73}}$ & 18.17$_{\textcolor{rise}{1.46}}$ & 44.67$_{\textcolor{rise}{1.33}}$ & 16.33$_{\textcolor{dec}{0.27}}$ & 36.17$_{\textcolor{dec}{2.44}}$ \\
            Mistral-7B   & 61.03$_{\textcolor{dec}{1.94}}$ & 72.62$_{\textcolor{dec}{1.07}}$ & 36.71$_{\textcolor{rise}{1.13}}$ & 70.05$_{\textcolor{dec}{0.85}}$ & 30.27$_{\textcolor{dec}{0.28}}$ & 58.62$_{\textcolor{dec}{1.87}}$ & 26.25$_{\textcolor{rise}{0.10}}$ & 51.05$_{\textcolor{rise}{0.04}}$ & 22.22$_{\textcolor{dec}{1.50}}$ & 43.00$_{\textcolor{dec}{2.62}}$ \\
			LLaMA3-8B    & 61.26$_{\textcolor{dec}{2.81}}$ & 72.12$_{\textcolor{dec}{1.62}}$ & 35.17$_{\textcolor{dec}{1.55}}$ & 70.83$_{\textcolor{rise}{0.01}}$ & 28.53$_{\textcolor{dec}{3.11}}$ & 59.00$_{\textcolor{dec}{1.88}}$ & 24.47$_{\textcolor{dec}{2.57}}$ & 51.38$_{\textcolor{dec}{0.24}}$ & 19.77$_{\textcolor{dec}{4.50}}$ & 42.46$_{\textcolor{dec}{3.62}}$ \\
Qwen2.5-7B   & 61.79$_{\textcolor{dec}{2.17}}$ & 73.33$_{\textcolor{dec}{1.31}}$ & 36.26$_{\textcolor{rise}{0.82}}$ & 69.59$_{\textcolor{dec}{0.77}}$ & 29.92$_{\textcolor{dec}{0.71}}$ & 57.72$_{\textcolor{dec}{2.80}}$ & 27.64$_{\textcolor{rise}{1.13}}$ & 53.66$_{\textcolor{rise}{0.80}}$ & 22.76$_{\textcolor{dec}{0.94}}$ & 45.04$_{\textcolor{dec}{1.99}}$ \\
\hdashline
ShieldLM-14B-Qwen  & 64.08$_{\textcolor{rise}{0.25}}$ & 74.48$_{\textcolor{rise}{1.03}}$ & 34.19$_{\textcolor{dec}{0.61}}$ & 69.86$_{\textcolor{dec}{0.37}}$ & 28.90$_{\textcolor{dec}{1.30}}$ & 58.30$_{\textcolor{dec}{1.51}}$ & 25.60$_{\textcolor{dec}{0.58}}$ & 51.69$_{\textcolor{rise}{0.45}}$ & 21.30$_{\textcolor{dec}{2.29}}$ & 43.60$_{\textcolor{dec}{1.98}}$ \\
ShieldGemma-9B   & 62.50$_{\textcolor{dec}{0.90}}$ & 74.35$_{\textcolor{rise}{0.04}}$ & 35.71$_{\textcolor{rise}{1.31}}$ & 71.10$_{\textcolor{dec}{0.01}}$ & 29.87$_{\textcolor{dec}{0.12}}$ & 60.71$_{\textcolor{dec}{0.80}}$ & 26.95$_{\textcolor{rise}{1.31}}$ & 55.03$_{\textcolor{rise}{2.33}}$ & 23.21$_{\textcolor{dec}{0.28}}$ & 46.10$_{\textcolor{dec}{1.04}}$ \\
    		\midrule
			\rowcolor{white} \multicolumn{11}{c}{\textit{LLM APIs (with Basic Prompt and 2 Examples)}} \\
			\midrule
LLaMA3.3-70B    & 29.22$_{\textcolor{rise}{2.90}}$ & 38.47$_{\textcolor{rise}{3.54}}$ & 14.24$_{\textcolor{rise}{0.75}}$ & 51.69$_{\textcolor{rise}{0.05}}$ & 6.75$_{\textcolor{rise}{0.79}}$ & 23.20$_{\textcolor{rise}{1.28}}$ & 6.75$_{\textcolor{rise}{1.83}}$ & 22.91$_{\textcolor{rise}{4.59}}$ & 4.26$_{\textcolor{rise}{1.04}}$ & 13.36$_{\textcolor{rise}{2.43}}$ \\
Qwen2.5-72B   & 45.58$_{\textcolor{rise}{4.64}}$ & 54.67$_{\textcolor{rise}{4.23}}$ & 22.15$_{\textcolor{rise}{1.05}}$ & 57.36$_{\textcolor{rise}{1.00}}$ & 16.77$_{\textcolor{rise}{1.11}}$ & 41.61$_{\textcolor{rise}{2.12}}$ & 12.42$_{\textcolor{dec}{0.06}}$ & 30.35$_{\textcolor{dec}{0.57}}$ & 8.83$_{\textcolor{rise}{0.09}}$ & 19.59$_{\textcolor{dec}{0.70}}$ \\
Gemini-2.5-Flash   & 46.10$_{\textcolor{rise}{3.29}}$ & 56.55$_{\textcolor{rise}{4.05}}$ & 21.19$_{\textcolor{rise}{0.85}}$ & 56.12$_{\textcolor{rise}{0.02}}$ & 16.18$_{\textcolor{rise}{1.26}}$ & 38.51$_{\textcolor{rise}{0.66}}$ & 13.17$_{\textcolor{rise}{1.02}}$ & 31.35$_{\textcolor{rise}{1.05}}$ & 9.74$_{\textcolor{rise}{0.05}}$ & 23.19$_{\textcolor{dec}{0.24}}$ \\
Claude-3.7-Sonnet   & 46.66$_{\textcolor{rise}{3.89}}$ & 58.14$_{\textcolor{rise}{4.97}}$ & 21.66$_{\textcolor{rise}{1.51}}$ & 59.30$_{\textcolor{rise}{1.04}}$ & 16.13$_{\textcolor{rise}{1.21}}$ & 41.86$_{\textcolor{rise}{1.80}}$ & 14.39$_{\textcolor{rise}{2.18}}$ & 38.08$_{\textcolor{rise}{4.82}}$ & 11.63$_{\textcolor{rise}{2.36}}$ & 28.63$_{\textcolor{rise}{3.83}}$ \\
GPT-4o    & 49.28$_{\textcolor{rise}{2.43}}$ & 61.30$_{\textcolor{rise}{3.11}}$ & 21.71$_{\textcolor{dec}{0.93}}$ & 59.66$_{\textcolor{dec}{2.75}}$ & 16.52$_{\textcolor{dec}{0.69}}$ & 46.55$_{\textcolor{rise}{0.14}}$ & 12.01$_{\textcolor{dec}{1.20}}$ & 33.72$_{\textcolor{dec}{1.96}}$ & 8.46$_{\textcolor{dec}{0.54}}$ & 22.80$_{\textcolor{dec}{0.54}}$ \\
DeepSeek-v3    & 52.69$_{\textcolor{rise}{4.53}}$ & 64.25$_{\textcolor{rise}{4.00}}$ & 23.79$_{\textcolor{rise}{1.00}}$ & 62.10$_{\textcolor{rise}{2.72}}$ & 19.22$_{\textcolor{rise}{0.54}}$ & 50.40$_{\textcolor{rise}{4.00}}$ & 16.13$_{\textcolor{rise}{1.18}}$ & 42.07$_{\textcolor{rise}{4.88}}$ & 11.69$_{\textcolor{rise}{0.21}}$ & 30.11$_{\textcolor{rise}{2.73}}$ \\
			\bottomrule
	\end{tabular}
\end{table*}

\subsection{Can LLMs understand posts containing coded hate terms?}

To evaluate the impact of coded hate terms on span-level Chinese hate speech detection models, we remove the posts without coded hate terms from the test set and obtain a subset of 502 posts containing coded hate terms for focused testing. By comparing the model's performance on this subset of data with its performance on the full test set, we can gain a clearer understanding of how hate slurs affect model performance. The experimental results are presented in Table \ref{slang}. The superscript numbers in the table indicate the difference in F1 score between the model's performance on the selected subset and its performance on the full test set. Green indicates a \textcolor{dec}{decrease} in F1 score on the subset compared to the full dataset, while red indicates an \textcolor{rise}{increase}.

\paragraph{Finetuned Models}
When evaluating performance on the subset containing coded hate terms, fine-tuned models generally exhibit a decline in both hard and soft matching metrics across most tasks. A notable exception is the hate speech triplet prediction task (\textit{T-A-H Tri.}), which shows an increase. This improvement can be attributed to two main factors: Firstly, posts with coded hate terms often convey a more overt hateful intent, in contrast to the more challenging implicit expressions present in the full dataset. This explicit nature facilitates the T-A-H Tri. task's ability to identify relevant relationships. Secondly, these findings highlight that the primary difficulties posed by coded hate terms for fine-tuned models lie in the precise extraction of targets and arguments, along with accurate classification of the targeted group. This suggests that without sufficiently comprehensive training data encompassing the complexities of coded hate, fine-tuned models struggle to adequately address these specific challenges.

\paragraph{LLM APIs}
In contrast, LLM APIs demonstrate improved performance on the test subset featuring coded hate terms. With the exception of the soft matching metric for argument identification, nearly all other metrics show an uplift. We propose two primary reasons for this: Firstly, coded hate terms frequently appear directly as targets, and these explicit target terms effectively mitigate the LLM APIs' inherent limitations in precise span boundary recognition. Secondly, deciphering coded hate terms often demands rich background knowledge, such as an understanding of cultural or historical contexts. This is precisely an area where LLM APIs, with their extensive pre-training, possess a significant advantage.

\paragraph{Comparison and Summary}

Fine-tuned models and LLM APIs exhibit distinct behaviors when encountering posts with coded hate terms. Fine-tuned models, while showing improved performance on \textit{T-A-H Tri.} task, often see F1 scores for target and argument span extraction decrease on these posts. This suggests an over-reliance on explicit cues for span identification, leading to struggles with nuanced boundaries in coded language. Conversely, LLM APIs demonstrate a relative performance improvement on such posts, likely leveraging their vast background knowledge to understand explicitly coded terms (which often act as targets) and partially compensating for their general limitations in precise span boundary detection. Despite this, their absolute performance on multi-component tasks remains significantly lower than fine-tuned models, highlighting a critical trade-off: fine-tuned models offer superior task-specific precision, whereas LLM APIs provide broader semantic and cultural understanding.

\section{Results on Coded Hate Terms}

\begin{table*}[t]
\caption{Results of LLMs on Coded Hate Term Identification (in \%).}
\centering
\begin{tabular}{lcccccc}
\toprule
\multirow{2}{*}{\textbf{Model}} 
& \multicolumn{2}{c}{\textbf{Overall}} 
& \multicolumn{2}{c}{\textbf{P-G Variation Terms}} 
& \multicolumn{2}{c}{\textbf{Hateful Semantic Terms}} \\
\cmidrule(lr){2-3} \cmidrule(lr){4-5} \cmidrule(lr){6-7} 
& \textbf{Hard} & \textbf{Soft} & \textbf{Hard} & \textbf{Soft} 
& \textbf{Hard} & \textbf{Soft} \\
\midrule
LLaMA3.3-70B   & 40.67 & 58.21 & 34.41 & 51.86 & 45.49 & 63.45 \\
Qwen2.5-72B    & 41.99 & 57.65 & 39.37 & 54.33 & 44.00 & 60.63 \\
Gemini-2.5-Flash & \underline{54.66} & \underline{66.50} & \underline{58.59} & \underline{69.86} & \underline{51.63} & \underline{64.31} \\
Claude-3.7-Sonnet & 38.25 & 55.40 & 38.32 & 55.24 & 38.20 & 55.90 \\
GPT-4o       & 31.12 & 41.57 & 27.23 & 38.32 & 34.11 & 44.55 \\  	
DeepSeek-v3  & \textbf{63.75} & \textbf{79.26} & \textbf{66.07} & \textbf{81.00} & \textbf{61.97} & \textbf{78.21} \\ 	
\bottomrule
\end{tabular}
\label{tab:task2_results}
\end{table*}

\begin{figure*}[h]
\centering
\includegraphics[scale=0.8]{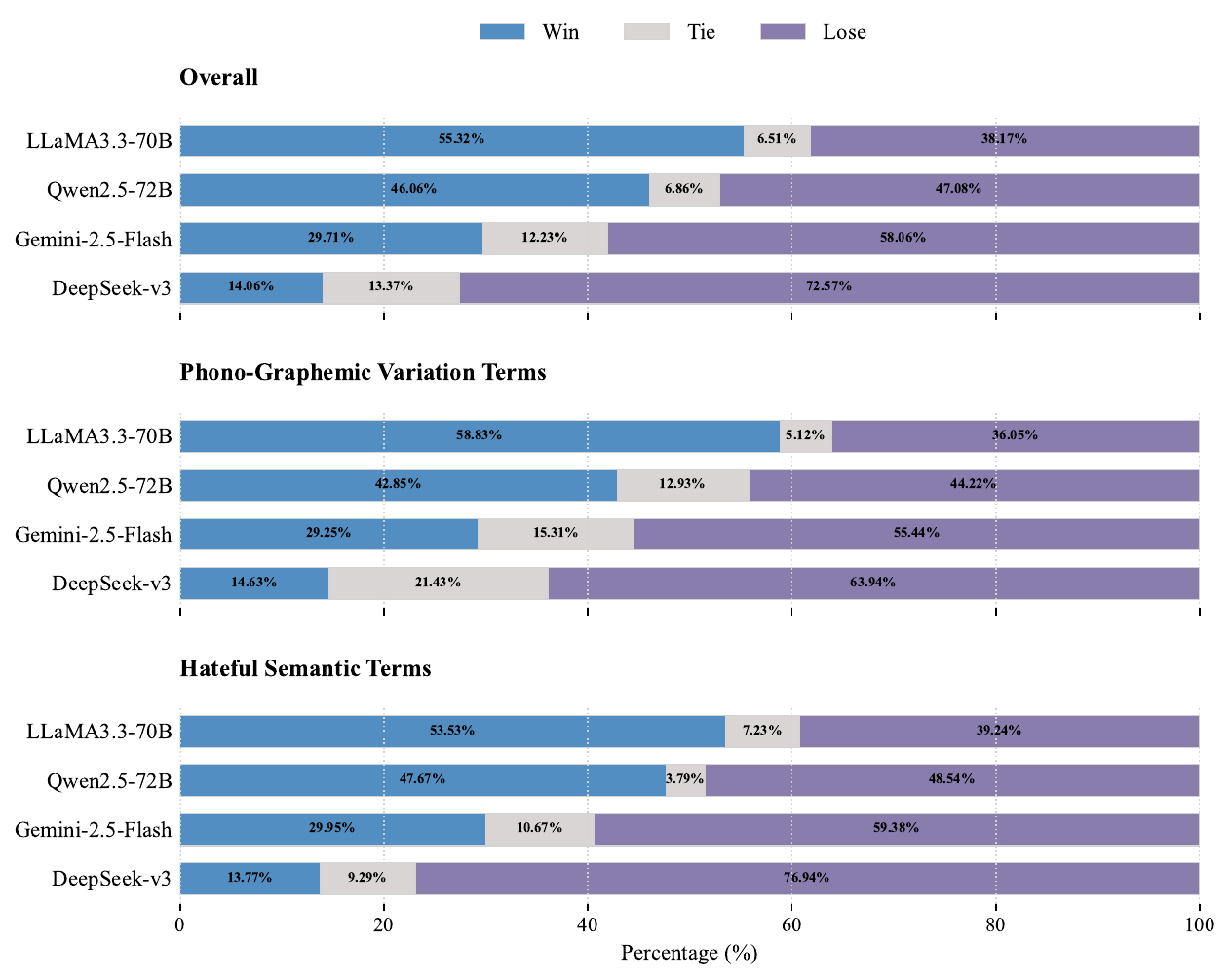} 
\caption{Results of pairwise comparison for coded hate term explanations.}
\label{case}
\end{figure*}

\subsection{Can LLMs identify Chinese coded hate terms?}

For \hyperref[task2]{Task 2}, the model's objective is to extract all coded hate terms from a given text. To prevent biases that could arise from manually incorporating definitions, we instead require the model to extract all hate terms
and match them against the annotated coded hate terms. Recall scores are adopted as the primary evaluation metric.

Our experiments reveal a significant performance disparity among leading LLMs in identifying Chinese coded hate terms. Overall, as detailed in Table \ref{tab:task2_results}, DeepSeek-v3 achieves remarkable hard and soft F1 scores of 63.75 and 79.26, respectively, indicating its superior ability. Gemini-2.5-Flash also demonstrates robust overall performance with scores of 54.66 (Hard) and 66.50 (Soft). In stark contrast, models such as GPT-4o (31.12 Hard, 41.57 Soft) and Claude-3.7-Sonnet (38.25 Hard, 55.40 Soft) struggled significantly, suggesting a critical weakness in interpreting these non-explicit forms of hate. A universal observation across all models is the considerable gap between soft and hard matching scores. While models can often approximate the location of a hateful term (Soft match), they generally lack the fine-grained precision to identify the exact span boundaries (Hard match), a persistent challenge in complex semantic extraction tasks.

For Phono-Graphemic Variation Terms, DeepSeek-v3 excels with scores of 66.07 (Hard) and 81.00 (Soft), and Gemini-2.5-Flash also performs strongly (58.59 Hard, 69.86 Soft). This indicates these models' advanced capacity not only to detect hate speech disguised through phonetic or orthographic variations but also for precise span localization within such complex expressions. GPT-4o's particularly low hard-matching score (27.23) on these terms underscores the challenge of interpreting such non-explicit forms of hate, which require multi-step reasoning to deconstruct linguistic disguises. Conversely, for Hateful Semantic Terms, whose offensiveness is rooted in cultural context, DeepSeek-v3 (61.97 Hard, 78.21 Soft) and Gemini-2.5-Flash (51.63 Hard, 64.31 Soft) maintain their strong performance. This suggests that their success points towards more advanced reasoning mechanisms that can generalize from learned patterns to new, unseen coded terms, moving beyond mere lexical recognition. Proficiency in identifying Chinese coded hate terms demands deep, context-aware semantic and cultural understanding, rather than just recognizing explicit terms from their training data.

\subsection{Can LLMs understand Chinese coded hate terms?}

For \hyperref[task3]{Task 3}, the model's objective is to explain the meaning of a coded hate term identified in a given text. Due to the labor-intensive and time-consuming nature of manually evaluating such explanations, we randomly sampled 100 data points for human annotation. To prevent pre-exposure bias, we recruited seven new annotators for this task. We then instructed GPT-4o \cite{gpt4o} to serve as an assessor, determining the winner between human and LLM explanations (allowing for ties) in Fig. \ref{case}.

LLMs demonstrate a clear hierarchy in their ability to explain Chinese coded hate terms, with performance significantly dictated by both their model architecture and the specific linguistic properties of the terms in question. Overall, DeepSeek-v3 demonstrates exceptional capacity, achieving a 72.57\% win rate against human explanations, indicating its strong ability to articulate nuanced meanings. Gemini-2.5-Flash also performs commendably with a 58.06\% win rate. This success suggests that certain LLMs, through their extensive pre-training on diverse texts, have effectively internalized complex cultural and social knowledge, enabling them to deconstruct and explain offensive connotations that might otherwise remain opaque. Conversely, LLaMA3.3-70B's substantial 55.32\% loss rate highlights that not all large models possess the same depth of understanding or the ability to synthesize this knowledge into coherent, high-quality explanations, underscoring significant variability in explanatory capabilities across powerful LLMs.

Across all tested models, explaining Phono-Graphemic Variation Terms proved significantly more challenging than explaining Hateful Semantic Terms. For instance, DeepSeek-v3's win rate dropped from an impressive 76.94\% for semantic terms to 63.84\% for disguised ones. This discrepancy suggests that deciphering the intent behind linguistically obfuscated terms (e.g., homophones, character manipulations) requires an additional, more complex layer of deconstructive reasoning. Models must first "decode" the phonetic or orthographic disguise before they can even begin to access the underlying hateful semantics. This multi-step inferential process appears to be a common bottleneck across architectures. In contrast, Hateful Semantic Terms, while culturally dependent, are often semantically more direct once their context is understood. The higher performance on these terms implies that leading LLMs excel when the primary challenge is semantic and socio-cultural interpretation rather than active linguistic reconstruction.


\section{Results on Two-stage Framework}

\begin{table*}[t]
\caption{Results of Two-Stage Fine-tuning for Offensive Language Detection with Coded Hate Term Lexicon (in \%).}
\centering
\begin{tabular}{lcccccccc}
\toprule
\multirow{2}{*}{\textbf{Model}} 
& \multicolumn{4}{c}{\textbf{COLD}} 
& \multicolumn{4}{c}{\textbf{CDial-Bias}} \\
\cmidrule(lr){2-5} \cmidrule(lr){6-9}
& \textbf{Acc.} & \textbf{Pre.} & \textbf{Rec.} & \textbf{F1} 
& \textbf{Acc.} & \textbf{MP} & \textbf{MR} & \textbf{F1} \\
\midrule
Qwen2.5-7B     & 70.30 & 79.80 & 50.87 & 62.13 & 60.57 & 56.00 & 45.11 & 44.60 \\
Only LoRA      & \underline{75.12} & \textbf{86.47} & 56.96 & 68.68 & \textbf{72.97} & \underline{62.46} & \underline{56.36} & \underline{58.27} \\
Only Lexicon   & 65.96 & 59.85 & \textbf{87.83} & \textbf{71.19} & 57.56 & 59.39 & 43.40 & 41.68 \\
Two Stage      & \textbf{76.16} & \underline{86.32} & \underline{59.67} & \underline{70.57} & \underline{72.60} & \textbf{63.81} & \textbf{58.00} & \textbf{60.04} \\
\bottomrule
\end{tabular}
\label{tab:comparison}
\end{table*}

\begin{table}[t]
\caption{Hyperparameter Settings for Two-stage Framework Finetuning.}
\centering
\begin{tabular}{lc}
\toprule
\textbf{Hyperparameters} & \textbf{Value} \\ 
\midrule
Epochs & 4 \\
Batch size & 1 \\
Learning rate & 5e-5 \\
Stage 2 learning rate & 2e-5 \\
Cutoff length & 1024 \\
Compute type & fp16 \\ 
Gradient accumulation & 8 \\
Maximum gradient norm & 1.0 \\ 
\bottomrule
\end{tabular}
\label{two_hyp}
\end{table}

To avoid label leakage and comprehensively evaluate our approach, we conducted experiments on two distinct Chinese datasets: COLD \cite{deng2022cold} and CDial-Bias \cite{zhou2022overview}. COLD is a widely recognized benchmark for binary classification of Chinese offensive language, while CDial-Bias focuses on three-class social bias detection in dialogues. For CDial-Bias, we specifically removed "Irrelevant data" labels to align with our evaluation objectives. The experimental results, as detailed in Table \ref{tab:comparison}, illustrate that incorporating a coded hate term lexicon significantly enhances offensive language detection in Chinese, especially when integrated into a two-stage fine-tuning framework.  "Only LoRA" refers to fine-tuning solely on the corresponding training dataset. "Only Lexicon" denotes fine-tuning exclusively on the Chinese coded hate term lexicon. The "Two Stage" approach involved sequential fine-tuning: first on the lexicon, then on the corresponding training dataset. Experiments are conducted on an RTX 4090D (24GB), with specific parameters detailed in Table \ref{two_hyp}.

\paragraph{Individual Strategies}

While "Only LoRA" effectively improves upon the Qwen2.5-7B baseline, confirming its general efficacy for task adaptation. Specifically, "Only LoRA" achieved an F1 score of 68.68 on COLD and 58.27 on Cdial-Bias, significantly surpassing the baseline's 62.13 and 44.60. "Only Lexicon" approach achieves a remarkably high recall of 87.83 on the COLD dataset, demonstrating the lexicon's power as a direct injection of domain-specific knowledge to identify hate terms. However, this broad coverage often comes at the cost of lower precision and accuracy, highlighting a susceptibility to false positives without further contextual refinement.

\paragraph{Two-Stage Framework}

The two-stage framework, which combines initial lexicon pre-tuning with subsequent dataset-specific fine-tuning, emerges as the optimal strategy. It consistently outperforms both "Only LoRA" and "Only Lexicon" by achieving the highest F1 scores on both COLD (70.57) and Cdial-Bias (60.04), alongside the top accuracy on COLD. This robust performance strongly validates the hypothesis that the lexicon provides a crucial foundational understanding of "coded hate terms". This understanding is then meticulously refined and contextualized by the subsequent training on specific datasets.

\paragraph{Comparison and Summary}
The two-stage framework's superior balance of precision and recall indicates that the initial lexicon exposure helps the model learn more robust, hate-specific feature representations. This, in turn, guides the later fine-tuning process towards a more comprehensive and accurate detection capability, effectively mitigating the lexicon-only approach's tendency for false positives while fully leveraging its high recall. This synergistic effect underscores the potential of systematically combining explicit domain knowledge (via lexicons) with implicit, data-driven learning (via LoRA on datasets) for achieving better performance in NLP tasks.

\section{Conclusion and Future Work}

With the advancement of hate speech detection, recent research has shifted from post-level detection to more fine-grained span-level detection. In this work, we focus on building resources for fine-grained Chinese hate speech detection. Firstly, we introduce \textsc{STATE ToxiCN}, the first span-level Chinese hate speech dataset, providing a crucial resource to evaluate models' deep semantic understanding of hateful content. Secondly, we conduct the first comprehensive study on Chinese coded hate terms, thoroughly assessing LLMs' ability to interpret their nuanced hate semantics. Our findings indicate a clear performance hierarchy among LLMs and reveal specific challenges in deciphering linguistically disguised terms. Thirdly, we propose and validate a two-stage fine-tuning framework that effectively integrates an annotated lexicon into detection models, demonstrably enhancing hate speech detection performance. This synergistic approach leverages the lexicon's direct knowledge injection to significantly improve both precision and recall. We hope that our resources and benchmarks will be valuable for researchers in this field.

Building on our foundational work, future research should explore how to effectively balance the broad background knowledge of large language models with the task-specific precision gained from fine-tuning for fine-grained analysis, mitigating potential capability loss. Concurrently, enhancing the adaptability and dynamism of our two-stage hate speech detection framework through evolving lexicons and adaptive learning mechanisms will be crucial to combat the constantly changing nature of coded hate speech. Furthermore, a significant future direction lies in expanding these methods into the multimodal domain, applying our findings to analyze hate speech across diverse media formats like images and videos.

\section*{Ethics Statement}

We adhere strictly to the data usage agreements of all public online social platforms and conduct thorough reviews to ensure that no user privacy information is included in our dataset. The opinions and findings reflected in the samples of our dataset do not represent the views of the authors, either explicitly or implicitly. We aim to ensure that the benefits of our proposed resources outweigh any potential risks. All resources are provided exclusively for scientific research purposes.

To minimize the psychological impact of evaluating harmful content, we have implemented a multi-faceted approach. This includes obtaining informed consent after thoroughly explaining the nature of the content they may encounter, carefully managing exposure by limiting weekly evaluation volumes, and empowering annotators to immediately cease work should they experience any discomfort. Furthermore, we proactively monitor their mental health through regular check-ins, ensuring a supportive and responsible working environment. 


\section*{Acknowledgments}
This research is supported by the Natural Science Foundation of China (No. 62376051, 61702080, 62366040), the Key R\&D Projects in Liaoning Province award numbers (2023JH26/10200015), the Fundamental Research Funds for the Central Universities (DUT24LAB123).


 

\bibliographystyle{IEEEtran}
\bibliography{paper}

\vspace{11pt}


\vfill
\end{CJK*}{UTF8}{gbsn}
\end{document}